\documentclass[10pt,twocolumn,letterpaper]{article}
\usepackage{iccv}

\usepackage{times}
\usepackage{epsfig}
\usepackage{graphicx}
\usepackage{amsmath}
\usepackage{amssymb}
\usepackage{multirow}
\usepackage{multicol}
\usepackage{caption}
\usepackage{subcaption}
\usepackage{booktabs}
\usepackage{enumitem}
\usepackage[table]{xcolor}

\DeclareMathOperator{\cU}{\mathcal{U}}
\DeclareMathOperator{\cHN}{\mathcal{HN}}
\DeclareMathOperator{\cN}{\mathcal{N}}

\usepackage[breaklinks=true,bookmarks=false]{hyperref}

\iccvfinalcopy 

\ificcvfinal\fi

\begin{document}

\title{RobustCLEVR: A Benchmark and Framework for Evaluating Robustness in Object-centric Learning}

\author{
Nathan Drenkow$^{1,2}$~~~~~Mathias Unberath$^1$ \\
$^1$The Johns Hopkins University \\
$^2$The Johns Hopkins University Applied Physics Laboratory
}

\maketitle

\ificcvfinal\thispagestyle{empty}\fi
\begin{abstract}
    Object-centric representation learning offers the potential to overcome limitations of image-level representations by explicitly parsing image scenes into their constituent components.  While image-level representations typically lack robustness to natural image corruptions, the robustness of object-centric methods remains largely untested. To address this gap, we present the RobustCLEVR benchmark dataset and evaluation framework. Our framework takes a novel approach to evaluating robustness by enabling the specification of causal dependencies in the image generation process grounded in expert knowledge and capable of producing a wide range of image corruptions unattainable in existing robustness evaluations.  Using our framework, we define several causal models of the image corruption process which explicitly encode assumptions about the causal relationships and distributions of each corruption type.  We generate dataset variants for each causal model on which we evaluate state-of-the-art object-centric methods. Overall, we find that object-centric methods are not inherently robust to image corruptions. Our causal evaluation approach exposes model sensitivities not observed using conventional evaluation processes, yielding greater insight into robustness differences across algorithms.  Lastly, while conventional robustness evaluations view corruptions as out-of-distribution, we use our causal framework to show that even training on in-distribution image corruptions does not guarantee increased model robustness. This work provides a step towards more concrete and substantiated understanding of model performance and deterioration under complex corruption processes of the real-world.\footnote{Data and code to be released soon}
\end{abstract}

\section{Introduction}
Common deep neural network (DNN) architectures have been shown to lack robustness to naturally-induced image-level degradation~\cite{Hendrycks2019-bm, Taori2020-qi, Djolonga2020-cq, Shankar2019-bv}.  In safety-critical scenarios, any reduction in model performance due to naturally-occurring corruptions poses a threat to system deployment.  While many proposed solutions exist for increasing the robustness of image-level representations~\cite{Zhang2017-oa, Cubuk2019-zh, Yun2019-av, Hendrycks2019-sc, Lee2020-sj, Chen2020-dr, Calian2021-mg, Modas2021-gw}, a measurable performance gap remains~\cite{Drenkow2021-zd, Taori2020-qi}.

\begin{figure}[t]
    \centering
    \includegraphics[width=0.75\linewidth]{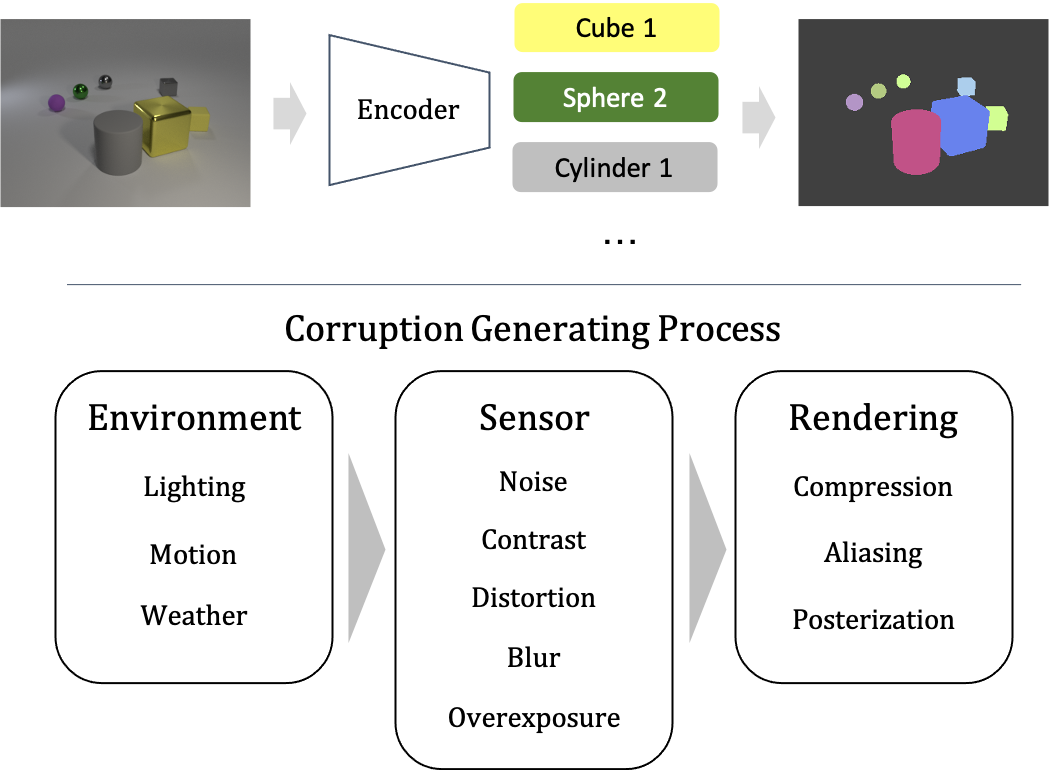}
    \caption{(Top) Object-centric methods explicitly parse scenes into constituent objects. (Bottom) 
    The corruption generating process involves with many causal factors with complex dependencies.}
    \label{fig:my_label}
\end{figure}

\begin{figure*}[t]
    \centering
    \includegraphics[width=0.6\linewidth]{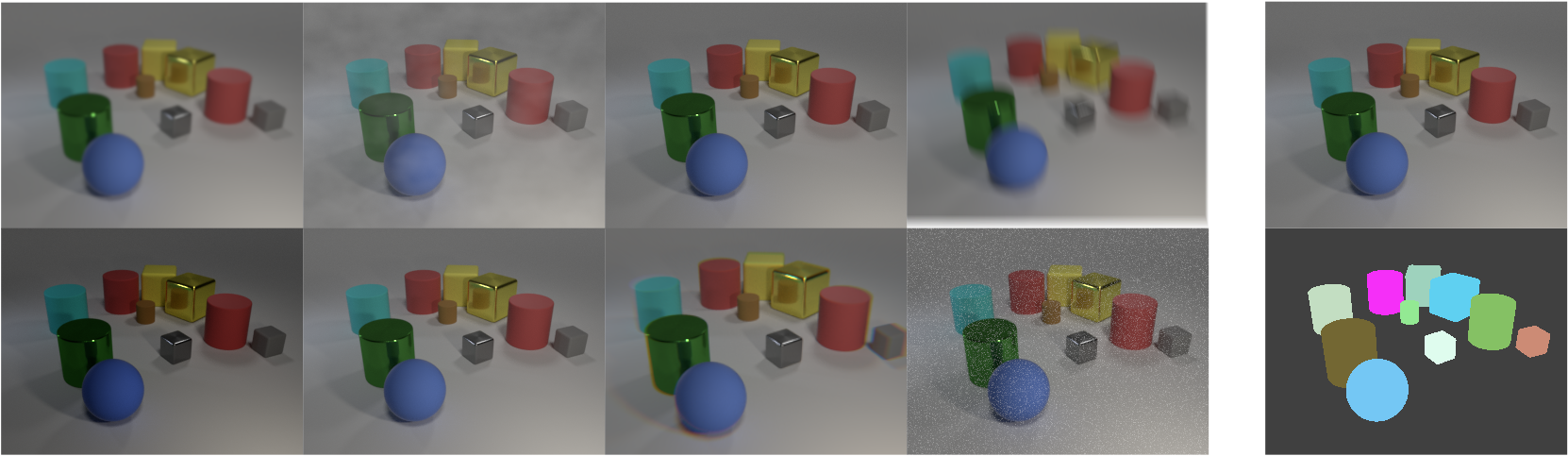}
    \caption{\textbf{RobustCLEVR image corruptions} rendered independently from left-to-right: (Top) Blur, cloud, defocus, displacement blur, (Bottom) Gamma, glare, lens distortion, noise.  Rightmost column is the clean image and ground truth mask.}
    \label{fig:iid-examples}
\end{figure*}
Recent advances in object-centric (OC) representation learning signal a paradigm shift towards methods that explicitly parse visual scenes as a precursor to downstream tasks.  These methods offer the potential to analyze complex scene geometries, support causal reasoning, and reduce the reliance of deep learning models on spurious image features and textures. Recent works have examined the use of OC representations for downstream tasks~\cite{Dittadi2021-wy} and action recognition~\cite{Zhang2022-pp} showing positive benefits of such representations over more traditional image-level features.  One hypothesis is that OC methods inherently learn scene-parsing mechanisms which are tied to stable features of the scene/objects and robust to naturally-induced image corruptions. However, quantitative proof of these desirable properties has not yet been obtained. We test this hypothesis by conducting the first analysis of the robustness of OC representation learning to non-adversarial, naturally-induced image corruptions.

\subsection{Background}
\label{sub:background}
\noindent\textbf{Robustness evaluation: } 
Robustness to common corruptions has been previously addressed in a number of other contexts~\cite{Hendrycks2019-bm, Laugros2019-mu, Kar2022-ae, Michaelis2019-iw, Maron2021-rr}.  However, prior work has made several strong limiting assumptions which we aim to address here. 
First, prior work has treated categories of image corruptions as IID, failing to account for causal relationships in the image generation process (e.g., low brightness causes longer exposure times or higher sensor sensitivity, resulting in motion artifacts or increased quantum noise, respectively). The lack of interactions leads to a less diverse set of image corruptions potentially decoupled from reality.

Second, corruption severity is often modeled heuristically without controlling for the impact on image quality and assuming all severities are equally likely.  Defining corruption severity on a discrete scale~\cite{Hendrycks2019-bm} often ignores the fact that severity is continuous in real-world conditions (e.g., blur due to motion depends on the velocity of the system or scene).  Furthermore, because DNNs are highly non-linear, performance change due to severity is also likely non-linear. Since corruption severity in the real world is often non-uniform, robustness evaluations should reconsider the nature of the assumed severity distribution.

Lastly, prior work typically assumes common corruptions are out-of-distribution (OOD), positing that they are not actually ``common'' (or even present) within the training sample distribution.  While this has benefits for assessing model performance on unseen conditions, it fails to consider the more likely scenario where the assumed distribution naturally contains image corruptions (even if rare) and the model may have access to corrupted samples during training.  Even in that case, the evaluation can still target specific corruption conditions while ensuring that the model has also been trained on data representative of the assumed ``true'' distribution.

\noindent\textbf{OC robustness evaluation: } 
OC representation learning is formulated as an unsupervised object discovery problem, so the absence of annotations (e.g., semantic labels, bounding boxes, object masks) forces models to learn only from the structure and imaging conditions inherent to the data distribution. To successfully develop OC methods that work on highly variable real-world data, robustness evaluations must be also able to account for a wider set of imaging conditions consistent with the image generation process. Training and evaluating the robustness of OC methods thus relies on a clear statement of the assumptions underlying the train and test distributions. 

\begin{figure*}[t]
    \centering
    \includegraphics[width=.95\linewidth]{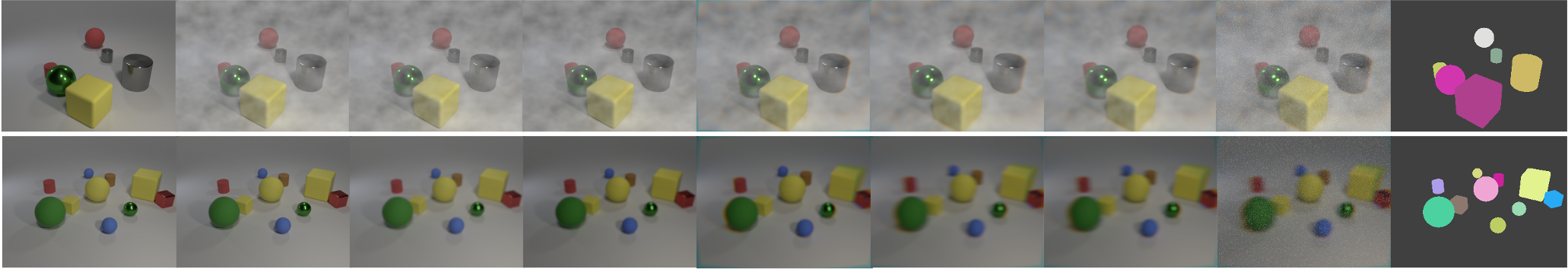}
    \caption{RobustCLEVR variant with causally-dependent corruptions. Rows are different samples from the same causal model and columns are images rendered at each node of the model. Corruptions are rendered according to order of the causal model from left to right: Clean, cloud, blur, gamma, lens distortion, displacement/motion blur, defocus blur, noise, ground truth.}
    \label{fig:non-iid-examples}
\end{figure*}

\noindent\textbf{Our approach: }
We unify and address limitations of prior work using a causal inference framework for robustness.  Knowledge of the image generation process enables the specification of causal graphs which explicitly capture  assumptions about the sources of and dependencies between image corruptions (described further in Section~\ref{sub:scm-dgp}).  We compare the robustness of OC methods using contrasting causal models of the data generating process (DGP) and show that assumptions about the causal model structure and its underlying distribution are critical for interpreting OC model robustness.  Lastly, we demonstrate that common corruption robustness of OC methods can also be interpreted as dealing with long-tail image distributions contrary to the more restrictive OOD assumption.

To investigate the robustness of OC methods, we developed the \textbf{RobustCLEVR} framework and dataset. We build off prior works which initially evaluated OC methods on CLEVR~\cite{Johnson2017-mg} and CLEVRTex~\cite{Karazija2022-zp}, datasets consisting of a collection of uncorrupted scenes composed of sets of simple objects with varying complexity of color, material, and texture properties.
We use our causal framework to generate multiple variants of RobustCLEVR with each investigating the effects of distributional assumptions on measured model robustness.  This benchmark and framework is a first of its kind for OC learning and provides a stepping stone towards realizing its potential on real-world data.

\subsection{Contributions}
\noindent Our work provides the following contributions:
\begin{itemize}[leftmargin=*]
    \setlength{\itemsep}{0pt}
    \item We perform a first evaluation of the robustness of object-centric representation learning methods to image-level corruptions in the conventional OOD setting.
    \item We introduce a causal framework for robustness which unifies common interpretations of robustness and grounds the data and evaluations in knowledge/assumptions of the image generation process.
    \item Using our causal framework, we develop the RobustCLEVR dataset containing CLEVR-like images under various forms of image degradation.  We provide multiple variants of the dataset generated from separate causal models capturing different assumptions about the image generation process.
    \item We run extensive evaluations and show that differences in assumptions about the underlying causal model have important implications for interpreting model robustness.
\end{itemize}

\section{Related Work}
\label{sec:related-work}
\noindent\textbf{Robustness benchmarks}
Robustness in deep learning for computer vision has been extensively studied outside of OC learning~\cite{Drenkow2021-zd}.  Several benchmarks have enabled systematic evaluation of robustness of deep learning methods with respect to image classification~\cite{Hendrycks2019-bm, Laugros2019-mu, Laugros2021-rq, Maron2021-rr}, object detection~\cite{Michaelis2019-iw}, instance segmentation~\cite{Altindis2021-uc}, and distribution shifts~\cite{Koh_undated-nq, Zhao2021-ek}. While challenging datasets for OC learning such as CLEVRTex~\cite{Karazija2021-ud} have helped push the boundaries of these methods, datasets for evaluating robustness to image corruption remains unexplored. 

\noindent\textbf{Object-centric methods} 
The problem of decomposing scenes into its constituent objects has been well-studied under a variety of labels including unsupervised object discovery, unsupervised semantic segmentation, and OC learning.  Early techniques~\cite{Eslami2016-pj, Kosiorek2018-at, Crawford2019-wl} processed images via a series of glimpses and learned generative models for scene construction by integrating representations extracted over multiple views.  More recent techniques learn models capable of generating full scenes from representations bound to individual objects. For generative~\cite{Burgess2019-ja, Engelcke2021-se, Jiang2020-ps, Lin2020-hp, Emami2021-cw, Greff2019-wm} and discriminative~\cite{Locatello2020-yv} methods, image reconstruction plays a crucial role in the learning objective.  

Beyond static scene images, multi-view and video datasets provide additional learning signals for unsupervised object discovery. Recent techniques~\cite{Singh2022-kp, Karazija2022-zp, Du_undated-wm, Bao2022-sm, Kipf2021-zo} have exploited object motion estimated via optical flow for improving OC representations.  In contrast, multi-view techniques~\cite{Nanbo2021-lw, Niemeyer2020-jv} take advantage of overlapping camera perspectives for improving scene decomposition. We focus on static scenes in this work and multi-view methods are instead candidates for future evaluation.

\noindent\textbf{Causal inference for robustness} 
Lastly, causal inference and computer vision research have become increasingly intertwined in recent years~\cite{Castro2020-wr, Drenkow2021-zd, Scholkopf2021-oq}.  Early works~\cite{Chalupka2014-jm, Lopez-Paz2016-sx} focused on causal feature learning and have since expanded to other vision tasks and domains~\cite{Mao2020-pv, Zhang2021-an, Qin2021-jb, Ilse2020-og}. Causal inference and robustness have also been explored in the context of adversarial~\cite{Zhang2021-an} and non-adversarial~\cite{Ding2022-oa, Moayeri2022-lc, Zhang2022-vv} conditions. 

\section{Methods}
\subsection{Structural causal models}
Structural causal models (SCM) consist of variables, their causal relationships, and their distributional assumptions, all of which describe an associated data generating process.  The DGP can be represented as a Directed Acyclic Graph (DAG) $\mathcal{G}$ consisting of variables ($\mathcal{V}$) as nodes and relationships ($\mathcal{E}$) as edges and where the output of a node is a function of its parents and an exogenous noise term (i.e., $v_i = f(pa_i, \epsilon_i)$ for node $i$). A joint distribution over all variables underlies the SCM which encodes their dependencies.

In computer vision, knowledge of the imaging domain and vision task provides a means for constructing such SCMs.  While full knowledge of $\mathcal{V}$, $\mathcal{E}$, and distributional information is generally not possible, plausible SCMs of the data generating process may still be constructed.  These SCMs encode expert knowledge and assumptions which can be verified through observational data. Alternatively, in simulated data, as in the case of CLEVR, we have full knowledge of the DGP including access to all variables and their underlying distributions. Critically, our framework allows us to leverage this access to fully specify \textit{arbitrary} graphical causal models of the DGP and then generate data in accordance with those models and their underlying distributions.

\subsection{Robustness}
\label{sub:robustness}
To date, the definition of robustness in computer vision has assumed many forms~\cite{Drenkow2021-zd} including (but not limited to) adversarial or worst-case behavior, out-of-distribution performance, and domain generalization. We aim to unify many of these interpretations via a causal framework.

First, we make a key distinction: the SCM of the data generating process describes our \textit{a priori} beliefs about the true data distribution, independent of any sampling of the data.  This provides a frame of reference for specifying robustness conditions such as when image corruptions are rare, due to distribution shift, or out-of-distribution. When evaluating robustness, we rely on this distinction in order to verify that the sampled training and evaluation datasets are consistent with our assumptions about the true underlying DGP.  

Formally, let $\mathcal{G}=(\mathcal{V}, \mathcal{E})$ be the structural causal model of a data generating process.  In the case of natural images, the nodes $\mathcal{V}=\{v_i\}$ represent variables such as the concepts of interest, distractor concepts, environmental conditions, and sensor properties. The model $\mathcal{G}$ induces a joint distribution $p_G(\{v_i\})$ over all variables $\mathcal{V}$ where $p(v_1, \dots, v_i) = \Pi_{j \leq i} p(v_j|pa(j))$ where $pa(j)$ are the parents of $j$ in $\mathcal{G}$.  

We consider common perspectives of robustness conditions in the context of structural causal models as follows.  
\begin{itemize}
    \setlength{\itemsep}{0pt}
    \item \textbf{Distribution shift} - Any shift in the marginal or conditional distributions of nodes in $\mathcal{V}$. 
    \item \textbf{Out-of-distribution (OOD)} - The case when test concepts or image conditions are not in the support of $p_\mathcal{G}(\{v_i\})$. This can be viewed as a special and extreme case of distribution shift.
    \item \textbf{Long-tail robustness} - Samples drawn from the DGP which are rare relative to the joint, marginal, and/or conditional distributions.
    \item \textbf{Adversarial} - Direct image manipulation performed via intervention (i.e., $do(X=X')$) 
\end{itemize}
This framework naturally allows for precise definition of the known/assumed robustness conditions as they relate to specific nodes of the DGP, which is in contrast to many common approaches which paint robustness in broad strokes. 
Conventional robustness evaluations are still included as a special case (i.e., IID corruptions assumed to be OOD) while more general evaluations may be implemented via soft/hard interventions on any subset of nodes in the SCM/DAG. These interventions measure the effects of specific types of corruptions on the image generating process by manipulating node values/distributions (independent of their parents) while maintaining downstream causal relationships.

\subsection{SCM of the Corruption Generating Process}
\label{sub:scm-dgp}
For studying the robustness of OC methods, we consider the case where image scenes composed of a finite set of objects are corrupted according to various imaging conditions.  Objects and scene geometry are first sampled independent of imaging conditions so that we can focus our attention on modeling the corruption generation process.  We define an SCM/DAG $\mathcal{G}=(\mathcal{C},\mathcal{E})$ where each $c_i$ applies a corruption to the already-constructed scene and edges $e_{ij}$ indicate dependencies between corruption types.  

For each corruption, we associate one (or more) severity parameter $\gamma_i$ such that for any image $x$, corruption $c$, and similarity metric $m(x,c(x; \gamma))$, we observe greater image degradation as $\gamma$ increases:
\begin{align*}
    \gamma_i > \gamma_j &\Rightarrow m(x, c(x; \gamma_i)) < m(x, c(x; \gamma_j)) \\
    &c(x; \gamma = 0) = x
\end{align*}     
Severity parameters are sampled from the causal model such that $\gamma_i = f_i(\gamma_{pa(i)}, \epsilon_i)$ where the causal mechanism $f_i$ is a function of $\gamma_{pa(i)}$, the severity parameters for the parents of node $i$, and $\epsilon_i$, a noise term. 

\begin{table*}[t!]
    \centering
    \caption{Mean Intersection over Union by model for IID-sampled corruptions. Rows within groups correspond to whether the corruption severity is sampled uniformly. Highlighted cells indicate the best performance in that column. }
    \resizebox{\textwidth}{!}{%
    \begin{tabular}{ll|rlrlrlrlrlrlrl|rl}
    \toprule
&  & \multicolumn{16}{c}{mIoU} \\
Model & Severity  & Blur &  & Clouds &  & Defocus &  & Gamma &  & Lens Distortion & & Motion Blur & & Noise &  & Clean &  \\
 \midrule
\multirow[c]{2}{*}{GENESISv2} & Non-uniform & 38.67 & $\pm$0.31 & 35.70 & $\pm$0.36 & 39.04 & $\pm$0.31 & 28.86 & $\pm$0.39 & 18.64 & $\pm$0.25 & 22.13 & $\pm$0.35 & 39.25 & $\pm$0.31 & 38.94 & $\pm$0.31 \\
 & Uniform & 39.24 & $\pm$0.31 & 38.35 & $\pm$0.32 & 38.93 & $\pm$0.31 & 26.01 & $\pm$0.40 & 26.27 & $\pm$0.28 & 30.27 & $\pm$0.33 & 39.54 & $\pm$0.31 & 39.00 & $\pm$0.31 \\
\cline{1-18}
\multirow[c]{2}{*}{GNM} & Non-uniform & 52.77 & $\pm$0.58 & 29.98 & $\pm$0.77 & 58.41 & $\pm$0.52 & {\cellcolor[HTML]{C0D6E4}} 51.75 & $\pm$0.75 & 27.13 & $\pm$0.52 & 24.88 & $\pm$0.65 & 56.71 & $\pm$0.53 & 61.32 & $\pm$0.50 \\
 & Uniform & 56.38 & $\pm$0.52 & 35.37 & $\pm$0.72 & 54.47 & $\pm$0.54 & 45.50 & $\pm$0.76 & {\cellcolor[HTML]{C0D6E4}} 43.14 & $\pm$0.50 & 40.08 & $\pm$0.63 & 58.01 & $\pm$0.51 & 61.01 & $\pm$0.50 \\
\cline{1-18}
\multirow[c]{2}{*}{IODINE} & Non-uniform & 63.75 & $\pm$0.45 & 32.83 & $\pm$0.96 & 66.60 & $\pm$0.42 & 27.84 & $\pm$0.77 & 26.83 & $\pm$0.46 & 30.78 & $\pm$0.64 & 65.84 & $\pm$0.45 & 66.20 & $\pm$0.40 \\
 & Uniform & 65.77 & $\pm$0.42 & 39.51 & $\pm$0.94 & 64.16 & $\pm$0.43 & 22.78 & $\pm$0.70 & 42.11 & $\pm$0.47 & 46.14 & $\pm$0.61 & 67.13 & $\pm$0.42 & 66.24 & $\pm$0.40 \\
\cline{1-18}
\multirow[c]{2}{*}{SPACE} & Non-uniform & 42.98 & $\pm$0.69 & 31.85 & $\pm$0.70 & 49.54 & $\pm$0.63 & 45.26 & $\pm$0.70 & 17.28 & $\pm$0.40 & 19.51 & $\pm$0.53 & 49.23 & $\pm$0.63 & 51.09 & $\pm$0.64 \\
 & Uniform & 48.37 & $\pm$0.62 & 37.65 & $\pm$0.67 & 45.37 & $\pm$0.63 & 42.95 & $\pm$0.72 & 28.36 & $\pm$0.48 & 30.10 & $\pm$0.62 & 49.85 & $\pm$0.63 & 50.95 & $\pm$0.63 \\
\cline{1-18}
\multirow[c]{2}{*}{SPAIR} & Non-uniform & 69.74 & $\pm$0.56 & 49.04 & $\pm$0.86 & {\cellcolor[HTML]{C0D6E4}} 71.86 & $\pm$0.56 & 30.85 & $\pm$0.91 & 23.60 & $\pm$0.49 & 31.19 & $\pm$0.68 & 71.13 & $\pm$0.57 & 72.99 & $\pm$0.57 \\
 & Uniform & {\cellcolor[HTML]{C0D6E4}} 71.22 & $\pm$0.55 & {\cellcolor[HTML]{C0D6E4}} 57.88 & $\pm$0.67 & 69.92 & $\pm$0.54 & 25.16 & $\pm$0.82 & 39.60 & $\pm$0.58 & {\cellcolor[HTML]{C0D6E4}} 46.99 & $\pm$0.68 & {\cellcolor[HTML]{C0D6E4}} 72.13 & $\pm$0.56 & {\cellcolor[HTML]{C0D6E4}} 73.04 & $\pm$0.56 \\
\cline{1-18}
\multirow[c]{2}{*}{eMORL} & Non-uniform & 18.05 & $\pm$0.24 & 14.23 & $\pm$0.24 & 18.47 & $\pm$0.25 & 14.62 & $\pm$0.22 & 12.05 & $\pm$0.17 & 12.47 & $\pm$0.19 & 17.86 & $\pm$0.24 & 18.54 & $\pm$0.25 \\
 & Uniform & 20.42 & $\pm$0.34 & 20.60 & $\pm$0.29 & 20.04 & $\pm$0.34 & 15.46 & $\pm$0.23 & 15.31 & $\pm$0.24 & 16.63 & $\pm$0.29 & 20.01 & $\pm$0.32 & 20.67 & $\pm$0.34 \\
\cline{1-18}
\bottomrule
\end{tabular}%
}
\label{tab:exp12-miou-bs}
\end{table*}

\begin{table*}[t!]
    \centering
    \caption{Mean Squared Error (MSE) by model for corruptions sampled IID.  Lower MSE indicates better recovery of the original clean image. Highlighted cells indicate the best performance in that column. }
    \resizebox{\textwidth}{!}{%
    \begin{tabular}{ll|rlrlrlrlrlrlrl|rl}
    \toprule
&  & \multicolumn{16}{c}{MSE} \\
Model & Severity  & Blur &  & Clouds &  & Defocus &  & Gamma &  & Lens Distortion & & Motion Blur & & Noise &  & Clean &  \\
 \midrule
\multirow[c]{2}{*}{GENESISv2} & Non-uniform & 58.60 & $\pm$2.06 & 193.36 & $\pm$6.33 & {\cellcolor[HTML]{C0D6E4}} 38.19 & $\pm$1.03 & {\cellcolor[HTML]{C0D6E4}} 216.07 & $\pm$7.13 & 599.75 & $\pm$11.79 & 402.55 & $\pm$9.64 & 38.10 & $\pm$1.06 & {\cellcolor[HTML]{C0D6E4}} 26.62 & $\pm$0.68 \\
 & Uniform & {\cellcolor[HTML]{C0D6E4}} 40.22 & $\pm$1.13 & {\cellcolor[HTML]{C0D6E4}} 131.23 & $\pm$3.88 & 50.49 & $\pm$1.55 & 259.92 & $\pm$6.41 & 324.83 & $\pm$7.27 & 230.88 & $\pm$6.14 & {\cellcolor[HTML]{C0D6E4}} 31.35 & $\pm$0.79 & 26.80 & $\pm$0.68 \\
\cline{1-18}
\multirow[c]{2}{*}{GNM} & Non-uniform & 117.60 & $\pm$2.63 & 288.40 & $\pm$7.31 & 96.39 & $\pm$1.93 & 875.12 & $\pm$36.45 & 598.87 & $\pm$12.20 & 390.67 & $\pm$8.81 & 103.53 & $\pm$2.08 & 87.10 & $\pm$1.79 \\
 & Uniform & 100.12 & $\pm$2.03 & 230.11 & $\pm$5.60 & 107.94 & $\pm$2.23 & 1182.75 & $\pm$36.29 & 319.91 & $\pm$7.05 & 246.39 & $\pm$5.55 & 95.93 & $\pm$1.92 & 87.42 & $\pm$1.78 \\
\cline{1-18}
\multirow[c]{2}{*}{IODINE} & Non-uniform & 76.52 & $\pm$2.16 & 455.67 & $\pm$13.47 & 58.05 & $\pm$1.51 & 1720.36 & $\pm$55.11 & 592.79 & $\pm$12.05 & 382.55 & $\pm$8.47 & 88.62 & $\pm$2.58 & 49.76 & $\pm$1.36 \\
 & Uniform & 60.74 & $\pm$1.55 & 381.64 & $\pm$12.13 & 69.67 & $\pm$1.81 & 2246.86 & $\pm$56.38 & 318.31 & $\pm$7.05 & 229.03 & $\pm$5.62 & 66.78 & $\pm$1.59 & 50.05 & $\pm$1.37 \\
\cline{1-18}
\multirow[c]{2}{*}{SPACE} & Non-uniform & 98.27 & $\pm$2.36 & 325.26 & $\pm$10.11 & 76.14 & $\pm$1.59 & 788.30 & $\pm$19.61 & 591.01 & $\pm$12.12 & 367.83 & $\pm$8.29 & 89.43 & $\pm$1.88 & 62.83 & $\pm$1.33 \\
 & Uniform & 80.95 & $\pm$1.74 & 221.29 & $\pm$5.85 & 90.13 & $\pm$2.04 & 950.58 & $\pm$19.44 & 319.09 & $\pm$7.00 & 220.79 & $\pm$5.19 & 78.04 & $\pm$1.59 & 63.13 & $\pm$1.36 \\
\cline{1-18}
\multirow[c]{2}{*}{SPAIR} & Non-uniform & 81.14 & $\pm$2.16 & 770.07 & $\pm$32.68 & 63.16 & $\pm$1.55 & 1797.57 & $\pm$59.94 & 623.68 & $\pm$12.88 & 408.85 & $\pm$9.15 & 90.24 & $\pm$2.48 & 53.18 & $\pm$1.41 \\
 & Uniform & 66.23 & $\pm$1.64 & 381.76 & $\pm$11.76 & 74.88 & $\pm$1.87 & 2377.58 & $\pm$61.43 & 334.74 & $\pm$7.39 & 247.47 & $\pm$6.13 & 69.51 & $\pm$1.68 & 53.60 & $\pm$1.43 \\
\cline{1-18}
\multirow[c]{2}{*}{eMORL} & Non-uniform & 62.58 & $\pm$2.04 & 1044.46 & $\pm$51.85 & 42.03 & $\pm$1.09 & 2014.28 & $\pm$58.91 & 609.37 & $\pm$12.68 & 371.51 & $\pm$8.61 & 85.20 & $\pm$2.92 & 32.11 & $\pm$0.84 \\
 & Uniform & 46.85 & $\pm$1.29 & 455.46 & $\pm$14.41 & 58.75 & $\pm$1.70 & 2662.18 & $\pm$63.08 & {\cellcolor[HTML]{C0D6E4}} 310.45 & $\pm$6.91 & {\cellcolor[HTML]{C0D6E4}} 211.35 & $\pm$5.40 & 51.65 & $\pm$1.24 & 31.78 & $\pm$0.91 \\
\cline{1-18}
\bottomrule
\end{tabular}%
}
\label{tab:exp12-mse-bs}
\end{table*}

\subsection{Generating RobustCLEVR}
\label{sub:rclevr}

The RobustCLEVR framework supports the definition of arbitrary SCMs/DAGs which capture various structural relationships and distributional assumptions regarding the data generating process and corresponding image corruptions.  Image corruptions are implemented via Blender workflows or ``recipes'' and are typically defined by one or a few corruption parameters (i.e., the $\gamma_i$ from Sec.~\ref{sub:scm-dgp}). While arbitrary corruption recipes may be defined using Blender to achieve a range of photorealistic effects, we implemented Gaussian blur, defocus blur, displacement/motion blur, gamma, clouds, white noise, glare, and lens distortion. 

For each image, the initial set of objects, their materials, and their placement in the scene are first sampled according to~\cite{Johnson2017-mg}. We then sample corruption parameters from the distribution defined by the SCM/DAG.  Using the Blender Python API, we apply the corruptions (given their sampled parameters) to the scene according to the ordering specified by the DAG.

Our framework evaluates robustness in two novel ways.  (1) Specification of the SCM/DAG allows for the generation of a wide range of \textit{unseen distortions} that may result from complex interdependencies/relationships between corruptions.  (2) Unlike prior works which consider corruption severity only at discrete and heuristic levels, samples from a DGP defined in our framework have corruption severities which vary continuously and consistent with the underlying distribution. Crucially, (1) and (2) enable better alignment with real world conditions where types of image corruptions rarely occur in pure isolation and their impact on image quality is continuously varying.

\section{Experiments and Results}
\label{sec:exp}

\textbf{Baseline Algorithms}
We evaluate pixel- and glimpse-based OC algorithms for all experiments.  For pixel-based methods we evaluate EfficientMORL~\cite{Emami2021-cw}, GENESISv2~\cite{Engelcke2021-se}, and IODINE~\cite{Greff2019-wm}.
For glimpse-based methods, GNM~\cite{Jiang2020-ps}, SPACE~\cite{Lin2020-hp}, and SPAIR~\cite{Crawford2019-wl} are evaluated. With the exception of Experiment 4 (Sec.~\ref{sub:exp4}), all models were trained on the public CLEVR training set and evaluated on the appropriate RobustCLEVR variants. We use code for baseline algorithms originally provided by~\cite{Karazija2021-ud}.

\textbf{Metrics} Consistent with prior work, we report performance on mean Intersection over Union (mIoU) and Mean Squared Error (MSE).  The mIoU metric measure the ability of the model to locate and isolate individual objects in the scene (i.e., object recovery) while MSE measures reconstruction quality (i.e., image recovery).  Metrics are computed relative to the uncorrupted images and the corresponding masks.  For each baseline in Experiments 1-3, we train a set of three models corresponding to different random seeds.  Due to significant variability in mIoU, we report metrics for the model+seed with the highest clean mIoU and obtain confidence intervals using 1000 bootstrap samples of predictions for each corruption. These results represent an upper bound on performance.  

\textbf{Dataset variants}  In each experiment, a test set is generated consisting of 10k distinct scenes.  For each scene, corruptions are rendered according to the parameters and ordering determined by the associated causal model. For eight corruption types and 10k scenes, this yields 80k images for evaluation per experiment.

\begin{figure*}[t]
    \centering
    \begin{subfigure}[b]{0.45\textwidth}
        \includegraphics[width=0.49\textwidth]{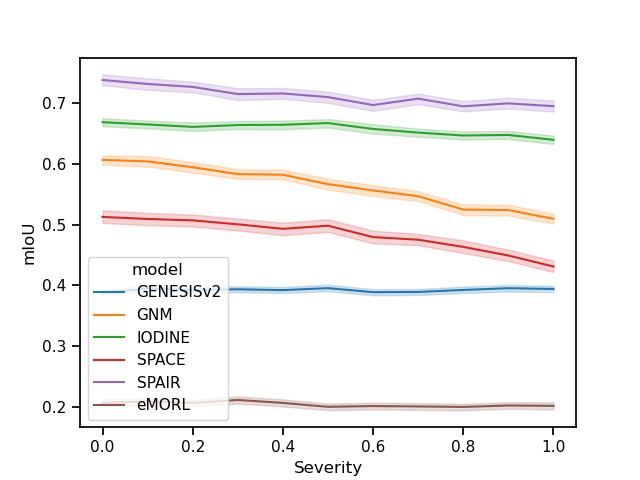}
        \includegraphics[width=0.49\textwidth]{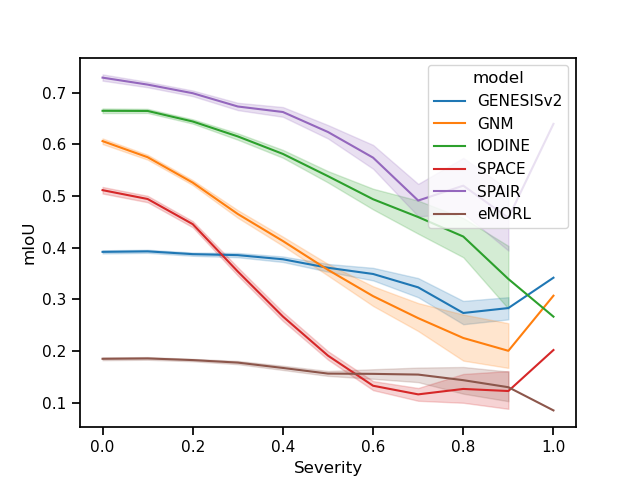}
        \caption{Blur corruption}
    \end{subfigure}
    \hfill
    \begin{subfigure}[b]{0.45\textwidth}
        \includegraphics[width=0.49\textwidth]{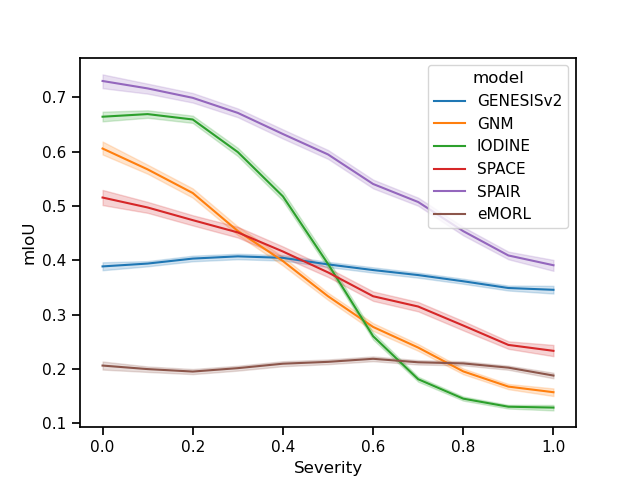}
        \includegraphics[width=0.49\textwidth]{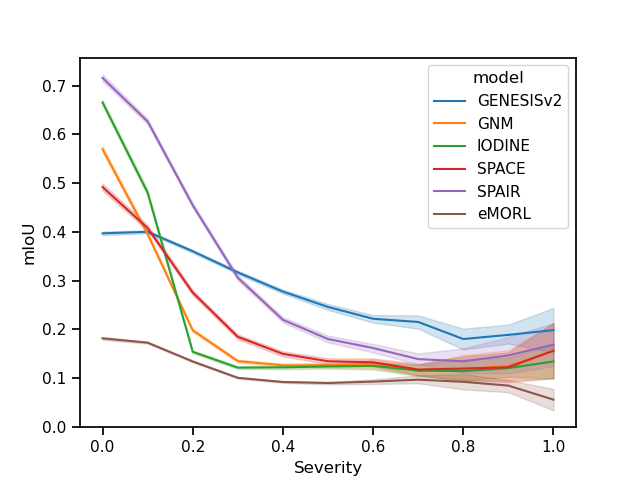}
        \caption{Cloud corruption}
    \end{subfigure}
    \caption{Object recovery (mIoU) as a function of normalized severity.  The severity is calculated by normalizing the sampled corruption parameter distribution to the interval $[0, 1]$ (with each panel normalized independently). For each corruption (panels (a), (b)), severity is sampled (left) uniformly and (right) non-uniformly.}
    \label{fig:miou-severity}
\end{figure*}

\subsection{Experiment 1: Independent corruptions, uniform severity}
\label{sub:exp1}
We first generate a RobustCLEVR variant where the causal model produces IID corruptions with uniform severity (i.e., corruption parameters $\gamma$ are sampled uniformly - See Appendix for distribution details).  This corresponds to the standard corruption evaluations where corruptions are OOD and independent with severity uniformly distributed.  Results of evaluating OC methods on this data are found in Tables~\ref{tab:exp12-miou-bs} and~\ref{tab:exp12-mse-bs}

The results indicate that the ability to recover underlying objects is largely tied to the distribution of corruption severity across the different corruption types. Figure~\ref{fig:miou-severity} shows how mIoU differs as a function of severity for each algorithm. For instance, for the cloud corruption, we see that SPAIR and IODINE report similar mIoU at low severity but SPAIR's performance degrades more gracefuly as severity increases. 

\begin{table*}[t!]
    \centering
    \caption{Mean Intersection over Union (mIoU) by model for corruptions sampled non-IID.  Corruption order in the table from left to right reflects the sampling order in the causal model. Higher mIoU indicates better recovery of the original clean image.}
    \resizebox{\linewidth}{!}{%
    \begin{tabular}{ll|rlrlrlrlrlrlrl|rl}
\toprule
 &  & \multicolumn{16}{c}{mIoU} \\
 Model & Severity & Clouds &  & Blur &  & Gamma &  & Lens Distortion &  & Motion Blur &  & Defocus &  & Noise &  & Clean &  \\
\midrule
\multirow[c]{2}{*}{GENESISv2} & Non-uniform & 38.75 & $\pm$0.30 & 38.88 & $\pm$0.30 & 36.88 & $\pm$0.31 & 34.32 & $\pm$0.31 & 31.79 & $\pm$0.29 & 31.80 & $\pm$0.29 & 32.28 & $\pm$0.29 & 38.86 & $\pm$0.30 \\
 & Uniform & 38.18 & $\pm$0.32 & 38.33 & $\pm$0.32 & 34.71 & $\pm$0.34 & 31.98 & $\pm$0.31 & 29.85 & $\pm$0.29 & 29.88 & $\pm$0.29 & 30.61 & $\pm$0.28 & 38.90 & $\pm$0.31 \\
\cline{1-18}
\multirow[c]{2}{*}{GNM} & Non-uniform & 55.02 & $\pm$0.73 & 52.95 & $\pm$0.71 & 55.54 & $\pm$0.70 & {\cellcolor[HTML]{C0D6E4}} 53.71 & $\pm$0.67 & {\cellcolor[HTML]{C0D6E4}} 48.78 & $\pm$0.65 & {\cellcolor[HTML]{C0D6E4}} 48.42 & $\pm$0.65 & 47.13 & $\pm$0.66 & 61.32 & $\pm$0.50 \\
 & Uniform & 53.87 & $\pm$0.77 & 50.81 & $\pm$0.74 & 53.29 & $\pm$0.75 & 51.18 & $\pm$0.71 & 47.04 & $\pm$0.69 & 45.56 & $\pm$0.67 & 44.68 & $\pm$0.69 & 60.99 & $\pm$0.51 \\
\cline{1-18}
\multirow[c]{2}{*}{IODINE} & Non-uniform & 59.87 & $\pm$0.74 & 59.56 & $\pm$0.75 & 49.88 & $\pm$0.85 & 46.49 & $\pm$0.78 & 41.34 & $\pm$0.66 & 41.16 & $\pm$0.67 & 41.75 & $\pm$0.69 & 66.56 & $\pm$0.41 \\
 & Uniform & 58.51 & $\pm$0.81 & 58.06 & $\pm$0.81 & 39.41 & $\pm$0.91 & 36.57 & $\pm$0.81 & 32.59 & $\pm$0.66 & 31.85 & $\pm$0.66 & 32.85 & $\pm$0.69 & 66.40 & $\pm$0.40 \\
\cline{1-18}
\multirow[c]{2}{*}{SPACE} & Non-uniform & 47.92 & $\pm$0.67 & 46.76 & $\pm$0.66 & 47.69 & $\pm$0.66 & 42.95 & $\pm$0.62 & 37.56 & $\pm$0.61 & 37.36 & $\pm$0.60 & 36.31 & $\pm$0.60 & 51.31 & $\pm$0.63 \\
 & Uniform & 46.49 & $\pm$0.72 & 44.78 & $\pm$0.70 & 45.76 & $\pm$0.69 & 40.89 & $\pm$0.64 & 36.51 & $\pm$0.63 & 35.17 & $\pm$0.61 & 33.74 & $\pm$0.60 & 51.07 & $\pm$0.62 \\
\cline{1-18}
\multirow[c]{2}{*}{SPAIR} & Non-uniform & {\cellcolor[HTML]{C0D6E4}} 69.60 & $\pm$0.66 & {\cellcolor[HTML]{C0D6E4}} 68.73 & $\pm$0.65 & {\cellcolor[HTML]{C0D6E4}} 58.70 & $\pm$0.89 & 53.24 & $\pm$0.82 & 46.53 & $\pm$0.69 & 46.37 & $\pm$0.70 & {\cellcolor[HTML]{C0D6E4}} 47.14 & $\pm$0.69 & {\cellcolor[HTML]{C0D6E4}} 73.28 & $\pm$0.57 \\
 & Uniform & 67.35 & $\pm$0.76 & 66.13 & $\pm$0.74 & 46.83 & $\pm$1.01 & 41.86 & $\pm$0.87 & 36.70 & $\pm$0.70 & 36.12 & $\pm$0.71 & 36.99 & $\pm$0.72 & 72.93 & $\pm$0.57 \\
\cline{1-18}
\multirow[c]{2}{*}{eMORL} & Non-uniform & 17.98 & $\pm$0.25 & 17.92 & $\pm$0.25 & 17.12 & $\pm$0.25 & 16.46 & $\pm$0.24 & 15.61 & $\pm$0.22 & 15.54 & $\pm$0.22 & 15.39 & $\pm$0.22 & 18.60 & $\pm$0.26 \\
 & Uniform & 17.59 & $\pm$0.26 & 17.48 & $\pm$0.26 & 15.78 & $\pm$0.24 & 15.22 & $\pm$0.22 & 14.57 & $\pm$0.21 & 14.33 & $\pm$0.21 & 14.19 & $\pm$0.21 & 18.60 & $\pm$0.25 \\
\cline{1-18}
\bottomrule
\end{tabular}
}%
    \label{tab:exp3-miou-bs}
\end{table*}

\begin{table*}[t!]
    \centering
    \caption{Mean Squared Error (MSE) by model for corruptions sampled non-IID.  Corruption order in the table from left to right reflects the sampling order in the causal model. Lower MSE indicates better recovery of the original clean image. }
    \resizebox{\linewidth}{!}{%
    \begin{tabular}{ll|rlrlrlrlrlrlrl|rl}
\toprule
 &  & \multicolumn{16}{c}{MSE} \\
 Model & Severity & Clouds &  & Blur &  & Gamma &  & Lens Distortion &  & Motion Blur &  & Defocus &  & Noise &  & Clean &  \\
\midrule
\multirow[c]{2}{*}{GENESISv2} & Non-uniform & {\cellcolor[HTML]{C0D6E4}} 52.49 & $\pm$2.70 & {\cellcolor[HTML]{C0D6E4}} 57.66 & $\pm$2.75 & {\cellcolor[HTML]{C0D6E4}} 80.43 & $\pm$3.15 & {\cellcolor[HTML]{C0D6E4}} 124.71 & $\pm$3.60 & {\cellcolor[HTML]{C0D6E4}} 171.34 & $\pm$3.90 & {\cellcolor[HTML]{C0D6E4}} 172.97 & $\pm$3.91 & {\cellcolor[HTML]{C0D6E4}} 166.59 & $\pm$3.90 & 26.62 & $\pm$0.69 \\
 & Uniform & 64.80 & $\pm$4.12 & 73.34 & $\pm$4.08 & 120.74 & $\pm$4.24 & 168.43 & $\pm$4.27 & 207.51 & $\pm$4.33 & 212.16 & $\pm$4.42 & 200.17 & $\pm$4.53 & {\cellcolor[HTML]{C0D6E4}} 26.62 & $\pm$0.67 \\
\cline{1-18}
\multirow[c]{2}{*}{GNM} & Non-uniform & 122.31 & $\pm$4.03 & 127.52 & $\pm$4.04 & 190.73 & $\pm$9.72 & 217.21 & $\pm$9.77 & 252.33 & $\pm$9.53 & 251.75 & $\pm$9.41 & 235.50 & $\pm$7.98 & 87.21 & $\pm$1.82 \\
 & Uniform & 133.39 & $\pm$5.09 & 141.30 & $\pm$5.04 & 289.28 & $\pm$12.18 & 317.00 & $\pm$11.81 & 346.80 & $\pm$11.37 & 346.07 & $\pm$11.25 & 302.66 & $\pm$9.02 & 87.25 & $\pm$1.77 \\
\cline{1-18}
\multirow[c]{2}{*}{IODINE} & Non-uniform & 131.41 & $\pm$8.26 & 137.09 & $\pm$8.27 & 382.47 & $\pm$22.97 & 417.21 & $\pm$22.81 & 457.00 & $\pm$22.06 & 458.36 & $\pm$22.04 & 419.13 & $\pm$19.75 & 50.02 & $\pm$1.39 \\
 & Uniform & 143.15 & $\pm$9.34 & 151.12 & $\pm$9.35 & 683.71 & $\pm$30.12 & 720.13 & $\pm$29.58 & 752.83 & $\pm$28.77 & 761.58 & $\pm$28.64 & 673.68 & $\pm$24.93 & 49.79 & $\pm$1.33 \\
\cline{1-18}
\multirow[c]{2}{*}{SPACE} & Non-uniform & 101.92 & $\pm$4.10 & 109.77 & $\pm$4.11 & 239.08 & $\pm$10.77 & 273.42 & $\pm$10.78 & 309.49 & $\pm$10.21 & 309.46 & $\pm$10.14 & 294.86 & $\pm$9.41 & 62.77 & $\pm$1.35 \\
 & Uniform & 123.10 & $\pm$6.43 & 134.88 & $\pm$6.36 & 401.79 & $\pm$14.03 & 439.68 & $\pm$13.84 & 469.27 & $\pm$13.06 & 468.05 & $\pm$12.73 & 440.53 & $\pm$11.80 & 63.13 & $\pm$1.35 \\
\cline{1-18}
\multirow[c]{2}{*}{SPAIR} & Non-uniform & 134.54 & $\pm$8.15 & 140.47 & $\pm$8.12 & 371.67 & $\pm$22.71 & 409.36 & $\pm$22.45 & 454.70 & $\pm$21.89 & 455.62 & $\pm$21.92 & 419.69 & $\pm$19.69 & 53.34 & $\pm$1.40 \\
 & Uniform & 216.09 & $\pm$18.82 & 224.80 & $\pm$18.73 & 721.41 & $\pm$31.21 & 759.17 & $\pm$30.44 & 797.32 & $\pm$29.55 & 803.72 & $\pm$29.72 & 726.93 & $\pm$27.35 & 53.23 & $\pm$1.38 \\
\cline{1-18}
\multirow[c]{2}{*}{eMORL} & Non-uniform & 147.29 & $\pm$11.24 & 153.64 & $\pm$11.29 & 469.62 & $\pm$26.98 & 503.31 & $\pm$26.72 & 545.94 & $\pm$26.04 & 544.75 & $\pm$25.84 & 502.95 & $\pm$23.45 & 32.26 & $\pm$0.84 \\
 & Uniform & 258.83 & $\pm$27.74 & 268.29 & $\pm$27.84 & 926.02 & $\pm$39.64 & 959.81 & $\pm$39.23 & 993.88 & $\pm$38.18 & 991.45 & $\pm$37.84 & 904.73 & $\pm$36.42 & 32.27 & $\pm$0.85 \\
\cline{1-18}
\bottomrule
\end{tabular}
}%
    \label{tab:exp3-mse-bs}
\end{table*}

\subsection{Experiment 2: Independent corruptions, non-uniform severity}
\label{sub:exp2}
We next examine the impact of independent corruptions with non-uniform severity.  A similar RobustCLEVR variant is generated with the same DAG as Experiment 1 but where the corruption parameter(s) for each node are sampled from non-uniform distributions.  Since for most parameters, monotonically increasing/decreasing the value of a corruption parameter corresponds to an increase in the severity of the corruption, the uniform distribution from Experiment~\ref{sub:exp1} is replaced with a Half-Normal distribution. This trades off a bias towards low-severity cases with the possibility of sampling higher severity cases from the distribution tails. Images from this variant are visualized in Figure~\ref{fig:iid-examples}.  Results of evaluating OC methods on this data are found in Tables~\ref{tab:exp12-miou-bs} and~\ref{tab:exp12-mse-bs}.

Results show that long-tailed severity distributions lead to measurable changes in absolute and relative values of mIoU across models.  For instance, performance generally improves for the Gamma corruption when the severity distribution shifts from uniform to non-uniform, whereas Lens Distortion or Motion Blur exhibit lower performance as a result of the long-tail. Furthermore, Figure~\ref{fig:miou-severity} also illustrates non-linear relationships between performance and severity. 

\subsection{Experiment 3: Dependent corruptions}
\label{sub:exp3}
An important benefit of the causal graph relative to the current standard robustness evaluation approach is the ability to describe causal relationships known or assumed to exist in the image domain of interest. As such, we next consider a more challenging RobustCLEVR variant where the underlying causal model follows a chain structure. Corruptions are linked sequentially and sampled corruption parameters are a function of the parameter values of their immediate parent. 

As in Experiment~\ref{sub:exp2}, we also consider the impact of distributional assumptions on the measured robustness.  We create an additional variant of the chain model with non-uniform severity distributions and evaluate the performance of OC methods on the data generated from this model as well.  While the causal model variants in this experiment no longer produce IID corruptions as in Experiment 1 and 2, the evaluation is still considered OOD since all models were trained on only clean CLEVR data. Results for Experiment 3 are found in Tables~\ref{tab:exp3-miou-bs} and~\ref{tab:exp3-mse-bs}.

While the chain DAG structure suggests that the total image corruption increases as images are sampled in sequence along the DAG, the causal mechanisms and distributions at each node also dictate how each corruption severity is sampled. For instance, this may lead to larger differences in performance from one corruption to the next in the model (e.g., Blur $\rightarrow$ Gamma vs. Defocus $\rightarrow$ Noise).

\subsection{Experiment 4: Long-tail Robustness}
\label{sub:exp4}
Lastly, many real world scenarios allow for the possibility that corrupted images are in-distribution (ID) but occur infrequently in the training set (either due to the rarity of the corruption in reality or due to sampling bias such as preferences by annotators for labeling clean images). We generate a RobustCLEVR variant which treats the corruptions as ID but rare.  As in Section~\ref{sub:exp2}, the causal model DAG is specified as a tree of depth 1 whereby all corruptions are mutually independent and severities are non-uniformly distributed (see Appendix for details).  For each OC method, we train two models, one on only clean data and one on data including corruptions with $p_{corr} = 0.01, p_{clean} = 1 - \sum_{i}p_{corr_i}$.  Each training dataset consists of 50k unique scenes.

All models are evaluated on a separate corrupted test set sampled from the same causal model used for training as well as the test set from Experiment 1 which contained IID corruptions with uniform severity.  These test sets correspond to the long-tail robustness and distribution shift cases described in Section~\ref{sub:robustness}. Results are shown in Table~\ref{tab:exp4}.

With all models (excluding SPACE), the inclusion of corrupted samples in the training set appears to generally decrease robustness for the corresponding model.  For models like GNM and GENESISv2, the performance differences are small whereas models like IODINE and eMORL often differ by $>10\%$ when corruptions are included/excluded from the training set.  These trends hold for evaluation on both the uniform and non-uniform distributions for severity. This is discussed in more detail in Section~\ref{sec:discussion}.

\begin{table*}[t]
    \centering
    \caption{Comparison of model performance when corruptions with non-uniform severity are in-distribution (clean + corrupt) and out of distribution (clean only).}
    \resizebox{0.85\linewidth}{!}{%
    \begin{tabular}{lll|ccccccc|c}
\toprule
 &  &  & \multicolumn{8}{c}{mIoU} \\
 Model & Train Distribution & Severity & Blur & Clouds & Defocus & Motion Blur & Gamma & Lens Distortion & Noise & Clean \\
\midrule
\multirow[c]{4}{*}{GENESISv2} & clean & Non-uniform & 0.218 & 0.201 & 0.224 & 0.137 & 0.204 & 0.130 & 0.224 & 0.225 \\
\cline{2-11}
 & clean + corrupt & Non-uniform & 0.191 & 0.185 & 0.193 & 0.147 & 0.196 & 0.121 & 0.189 & 0.194 \\
\cline{2-11}
 & clean & Uniform & 0.224 & 0.217 & 0.221 & 0.179 & 0.200 & 0.167 & 0.226 & 0.225 \\
\cline{2-11}
 & clean + corrupt & Uniform & 0.193 & 0.188 & 0.192 & 0.167 & 0.196 & 0.153 & 0.192 & 0.195 \\
\cline{1-11} \cline{2-11}
\multirow[c]{4}{*}{GNM} & clean & Non-uniform & 0.466 & 0.243 & 0.524 & 0.226 & 0.493 & 0.247 & 0.489 & 0.550 \\
\cline{2-11}
 & clean + corrupt & Non-uniform & 0.456 & 0.231 & 0.515 & 0.221 & 0.367 & 0.246 & 0.487 & 0.543 \\
\cline{2-11}
 & clean & Uniform & 0.503 & 0.279 & 0.485 & 0.359 & 0.423 & 0.390 & 0.512 & 0.548 \\
\cline{2-11}
 & clean + corrupt & Uniform & 0.494 & 0.262 & 0.475 & 0.352 & 0.297 & 0.387 & 0.504 & 0.540 \\
\cline{1-11} \cline{2-11}
\multirow[c]{4}{*}{IODINE} & clean & Non-uniform & 0.627 & 0.303 & 0.651 & 0.310 & 0.275 & 0.262 & 0.628 & 0.647 \\
\cline{2-11}
 & clean + corrupt & Non-uniform & 0.274 & 0.210 & 0.284 & 0.180 & 0.290 & 0.155 & 0.277 & 0.288 \\
\cline{2-11}
 & clean & Uniform & 0.645 & 0.356 & 0.635 & {\cellcolor[HTML]{C0D6E4}} 0.463 & 0.230 & 0.412 & 0.649 & 0.649 \\
\cline{2-11}
 & clean + corrupt & Uniform & 0.283 & 0.233 & 0.279 & 0.228 & 0.275 & 0.207 & 0.281 & 0.288 \\
\cline{1-11} \cline{2-11}
\multirow[c]{4}{*}{SPACE} & clean & Non-uniform & 0.123 & 0.123 & 0.123 & 0.123 & 0.123 & 0.123 & 0.123 & 0.123 \\
\cline{2-11}
 & clean + corrupt & Non-uniform & 0.664 & 0.386 & {\cellcolor[HTML]{C0D6E4}} 0.717 & 0.273 & 0.617 & 0.255 & 0.708 & {\cellcolor[HTML]{C0D6E4}} 0.732 \\
\cline{2-11}
 & clean & Uniform & 0.123 & 0.123 & 0.123 & 0.123 & 0.123 & 0.123 & 0.123 & 0.123 \\
\cline{2-11}
 & clean + corrupt & Uniform & {\cellcolor[HTML]{C0D6E4}} 0.704 & 0.472 & 0.683 & 0.463 & 0.554 & {\cellcolor[HTML]{C0D6E4}} 0.434 & {\cellcolor[HTML]{C0D6E4}} 0.716 & 0.730 \\
\cline{1-11} \cline{2-11}
\multirow[c]{4}{*}{SPAIR} & clean & Non-uniform & 0.685 & 0.475 & 0.706 & 0.309 & 0.290 & 0.232 & 0.693 & 0.716 \\
\cline{2-11}
 & clean + corrupt & Non-uniform & 0.682 & 0.578 & 0.700 & 0.315 & {\cellcolor[HTML]{C0D6E4}} 0.624 & 0.230 & 0.697 & 0.708 \\
\cline{2-11}
 & clean & Uniform & 0.701 & 0.559 & 0.688 & 0.463 & 0.241 & 0.389 & 0.705 & 0.716 \\
\cline{2-11}
 & clean + corrupt & Uniform & 0.694 & {\cellcolor[HTML]{C0D6E4}} 0.640 & 0.682 & 0.463 & 0.615 & 0.388 & 0.703 & 0.707 \\
\cline{1-11} \cline{2-11}
\multirow[c]{4}{*}{eMORL} & clean & Non-uniform & 0.397 & 0.318 & 0.406 & 0.226 & 0.204 & 0.205 & 0.421 & 0.411 \\
\cline{2-11}
 & clean + corrupt & Non-uniform & 0.192 & 0.170 & 0.196 & 0.130 & 0.219 & 0.120 & 0.184 & 0.197 \\
\cline{2-11}
 & clean & Uniform & 0.405 & 0.384 & 0.400 & 0.313 & 0.181 & 0.283 & 0.420 & 0.410 \\
\cline{2-11}
 & clean + corrupt & Uniform & 0.197 & 0.179 & 0.195 & 0.161 & 0.204 & 0.143 & 0.185 & 0.199 \\
\cline{1-11} \cline{2-11}
\bottomrule
\end{tabular}%
    }
    \label{tab:exp4}
\end{table*}

\section{Discussion}
\label{sec:discussion}
The experiments in Section~\ref{sec:exp} suggest that OC methods are not immune to image corruptions.  While it is not surprising that performance degradation would occur in these cases, the sensitivity to low-severity corruptions suggests that OC models are not inherently more robust than non-OC techniques. We attribute much of this finding to the use of image reconstruction as a common component of the learning objective for these models. For generative methods, this is due to the log likelihood term in the ELBO objective while discriminative methods like Slot Attention use MSE directly. Consistent with results on CLEVRTex~\cite{Karazija2021-ud}, models which produce lower MSE (i.e., better image recovery) also tend to produce lower mIoU (i.e., object recovery). The use of image reconstruction by OC models during learning may encourage the latent representations to encode nuisance or appearance factors not critical to scene parsing. The result of this learning strategy is poor object recovery when those same nuisance factors are modified or corrupted as in a robustness scenario.

We also find that the structure and corresponding distribution of the underlying data generating process matters in assessing model robustness.   We observe measurable performance differences as a result of changing causal and distributional assumptions. For instance, considering two top models from Experiments 1-3, GNM and SPAIR, we observe differences in relative mIoU performance on the same set of corruptions drawn from the IID (Table~\ref{tab:exp12-miou-bs}) and non-IID causal models (Table~\ref{tab:exp3-miou-bs}).  While we expect the mIoU to change for each model as a result of the distribution shift, the disparity in mIoU between the two models for any given corruption is not constant between the IID and non-IID scenarios.  When causal models are defined to approximate specific real-world distributions, measuring such performance differences may be critical to understanding and predicting model behavior in the wild.

Lastly, the results of Experiment 4 indicate that robustness is not a purely OOD problem.  The inclusion of corrupted data as rare samples in the training distribution has a negative impact on robustness for many of the models.  This warrants further research as it contradicts existing findings for robustness in supervised, discriminative models where data augmentation with heavy corruption or other image transformations yields significant gains in robustness to common corruptions~\cite{Hendrycks2019-sc, Cubuk2019-zh, Modas2021-gw, Lee2020-sj, Chen2020-dr, Yun2019-av, Zhang2017-oa}. One possible explanation is that the corrupted images (while rare in training), simply provide less informative signal about the scene geometry and object properties.  Alternatively, when the training sample size is fixed, the inclusion of these corrupted images also means that fewer clean images are also available for learning.  
When corrupted images are in distribution, OC models with image reconstruction objectives may be increasingly incentivized to reconstruct low level corruptions which have no bearing on object recovery. So while OC methods aim to represent objects explicitly with less reliance on textures and other spurious image patterns, the reconstruction objective may unintentionally impose a barrier to success.

\paragraph{Limitations}
We note several limitations of this work to be addressed in future research.  First, defining causal models of the image/corruption generating process is not trivial and we make no claim that our RobustCLEVR corruption variants model the ``true'' causal mechanisms or distributions for real-world image corruptions. We also acknowledge that the full space of possible causal model graphs, mechanisms, and distributions is intractable to evaluate. Nonetheless, we evaluate two contrasting causal models which are sufficient to successfully demonstrate that OC model performance is highly dependent on the SCM and underlying data distribution. We also did not explore causal dependencies between properties of the scene and the occurrence of corruptions (e.g., the presence of dark settings only for specific objects). However, our variants instead intend to capture a wide range of image distortions independent of the scene composition with the purpose of more broadly testing OC methods beyond the conventional IID case.  Further, the corruptions in RobustCLEVR are applied late in the rendering pipeline which limits their overall realism. That said, CLEVR-like scenes are considerably simpler than real world data and the lack of robustness on RobustCLEVR images provides a useful check prior to testing on more complex scenes. Lastly, metrics are computed relative to ground truth image masks and clean images, yet in severe cases, corruptions will prevent OC methods from fully recovering the original objects/image.  While this may make it difficult to estimate the true upper bound on performance, this does not prevent relative comparisons between models.

\section{Conclusion}
In light of recent advances in object-centric learning, we present the first benchmark dataset for evaluating robustness to image corruptions. To thoroughly test robustness, we adopt a causal model framework whereby assumptions about the corruption generating process can be explicitly implemented and compared. We evaluate a set of state-of-the-art OC methods on data generated from causal models encoding various assumptions about the corruption generating process. We find that OC models are not robust to corruptions and further demonstrate through our causal model framework that distributional assumptions matter when comparing model robustness. While our results indicate that OC models are not implicitly robust to a range of natural image corruptions, object-centric learning still holds great promise for achieving robust models in the future.

{\small
\bibliographystyle{ieee_fullname}
\bibliography{egbib}
}

\clearpage
\section*{Appendix}
\section{Severity distributions for independent corruptions}
In the case of independent corruptions, two variants of the SCM were created which only differed by the distributions over corruption parameters.  Each parameter contributes directly to corruption severity and in all cases, larger parameter values produce larger corruptions.

\begin{table}[h!]
    \centering
    \caption{Corruption parameter distributions for the independent corruptions SCM}
    \resizebox{\columnwidth}{!}{%
    \begin{tabular}{c|c|c|c}
        \toprule
        & & \multicolumn{2}{c}{Distribution} \\
        \midrule
        Corruption & Parameter & Uniform & Non-uniform \\
        \midrule
        Gamma & $\gamma$ & $\cU(1, 3)$ & $\cHN(1)$ \\
        Blur & $\sigma$ & $\cU(1, 11)$ & $\cHN(11)$ \\
        Defocus & $z$, $f_{stop}$ & $\cU(1, 10)$, $\cU(64, 128)$ & $\cHN(3), \cU(64, 128)$\\
        Lens distort & distort, disperse & $\cU(0, 0.1), \cU(0, 0.5)$ & $\cHN(0.3), \cHN(0.7)$ \\
        Directional blur & distance & $\cU(0, 0.1)$ & $\cHN(0.2)$ \\
        Noise & scale & $\cU(0, 0.25)$ & $\cHN(0.25)$ \\
        Clouds & factor & $\cU(0, 0.3)$ & $\cHN(0.3)$ \\
        Glare & mix & $\cU(-0.5, 0.5)$ & $\cN(0, 0.5)$ \\
        \bottomrule
    \end{tabular}%
    }
    \label{tab:exp1-2-distributions}
\end{table}

\section{Causal model for dependent corruptions}
\begin{figure}[h]
    \centering
    \includegraphics[width=\columnwidth]{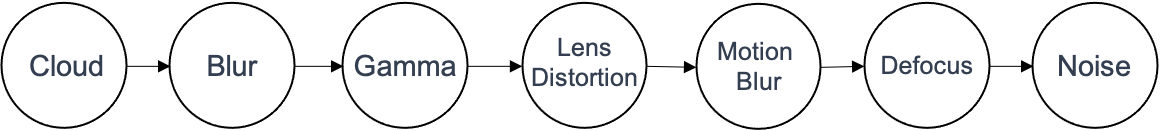}
    \caption{Causal model for non-IID image corruptions}
    \label{fig:exp3-scm}
\end{figure}

The causal model used for Experiment 3 follows the structure in Figure~\ref{fig:exp3-scm}. Each corruption parameter $\gamma_i$ is a function of its parents and the exogenous noise $\epsilon_i$ (i.e., $\gamma_i = f(\gamma_{pa(i)}, \epsilon_i)$).

\noindent The structural equations for this model are as follows:

\noindent\textbf{Clouds:}
\begin{align*}
    \epsilon &\sim \mathcal{U}(0,1) \\
    factor(\epsilon) &= \begin{cases}
        0 &\epsilon < 0.75 \\
        x \sim \mathcal{HN}(0.3) &else
    \end{cases}
\end{align*}

\noindent\textbf{Blur:}
\begin{align*}
    \sigma(factor) = \begin{cases}
        1 &factor > 0.2 \\
        k\in \{1..9\},~p(k) = 1/9 &else
    \end{cases}
\end{align*}

\noindent\textbf{Gamma:}
\begin{align*}
    \epsilon &\sim \mathcal{U}(0, 1) \\
    \gamma(k, \epsilon) &= \begin{cases}
        0.1 \cdot \epsilon + 1 &k \leq 3 \\
        \epsilon + 1 &k > 3 \\
    \end{cases}
\end{align*}

\noindent\textbf{Lens distortion:}
\begin{align*}
    \epsilon &\sim \mathcal{U}(0, 1) \\
    distort(\gamma, \epsilon) &= \begin{cases}
        0.05 \cdot \epsilon &\gamma > 1.2 \\
        0.5 \cdot \epsilon &1.0 < \gamma \leq 1.2 \\
    \end{cases}
\end{align*}

\noindent\textbf{Displacement/motion blur:}
\begin{align*}
    \epsilon_z, \epsilon_d &\sim \mathcal{U}(0,1) \\
    zoom(distort, \epsilon_z) &= 0.1 \cdot \epsilon_z \\
    distance(distort, \epsilon_d) &= 0.05 \cdot \epsilon_d 
\end{align*}

\noindent\textbf{Defocus blur:}
\begin{align*}
    z(zoom, distance) &= \begin{cases}
        x \in \{1..10\},~p(x)=1/10; \\
         \qquad (zoom = 0) \vee (distance = 0) \\
        1 \qquad else
    \end{cases} \\
    f_stop(zoom, distance) &= \begin{cases}
        x \in \{64..128\},~p(x)=1/64; \\
            \qquad (zoom = 0) \vee (distance = 0) \\
        128; \qquad else
    \end{cases}
\end{align*}

\noindent\textbf{Noise:}
\begin{align*}
    \epsilon &\sim \mathcal{U}(0, 1) \\
    \sigma_n(z, f_stop, \epsilon) &= \begin{cases}
        0.2 \cdot \epsilon &(z > 4) \vee (f_stop > 100) \\
        0.05 \cdot \epsilon &else
    \end{cases}
\end{align*}

This structural causal model is meant to induce wide variability in the image corruptions in contrast to the IID model.  The structural dependencies were specified to produce more visually complex corruptions. A comparison of Experiment 1-3 illustrates how OC model performance varies significantly with changes in the corruption generating process.

\section{Object recovery vs. Severity by Corruption}
The following figures compare the per-corruption performance of models as a function of normalized severity.  In each case, normalized severity is determined by the range of the sampled corruption parameters. The top panel in each figure corresponds to the case where corruptions are sampled IID (Experiment 1) and the bottom panel corresponds to the non-IID case (Experiment 3). 

\begin{figure}[h!]
    \centering
    \includegraphics[width=\linewidth]{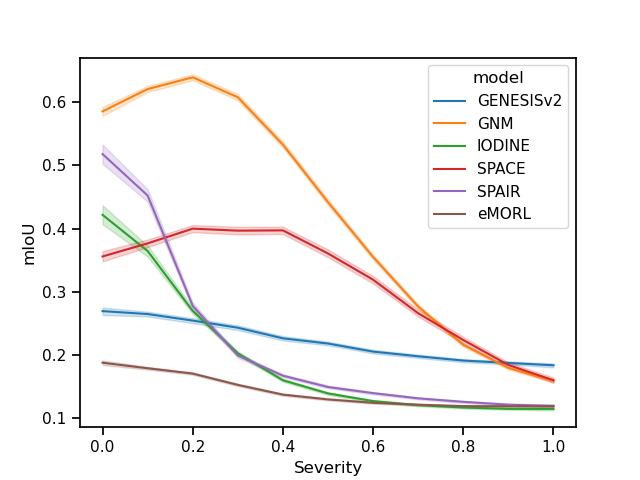} \\
    \includegraphics[width=\linewidth]{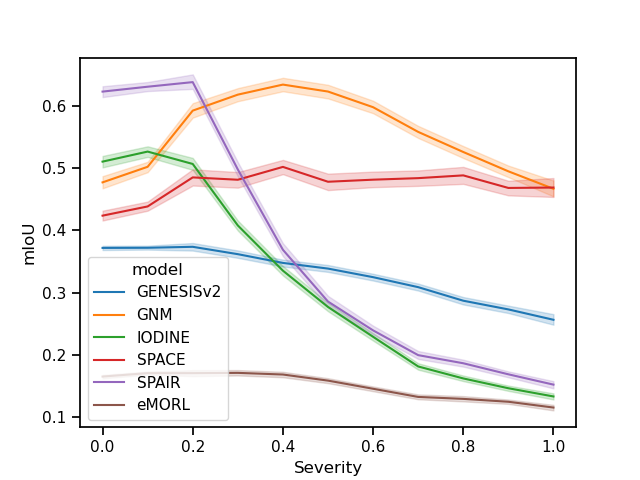}
    \caption{Gamma corruption}
    \label{fig:gamma}
\end{figure}

The figures show that mIoU (object recovery) as a function of severity may vary significantly across algorithms.  For instance, in Figure~\ref{fig:gamma}, both IODINE and SPAIR produce the highest clean performance, but mIoU drops rapidly with severity which is in stark contrast to the other methods analyzed.

\begin{figure}[h!]
    \centering
    \includegraphics[width=\linewidth]{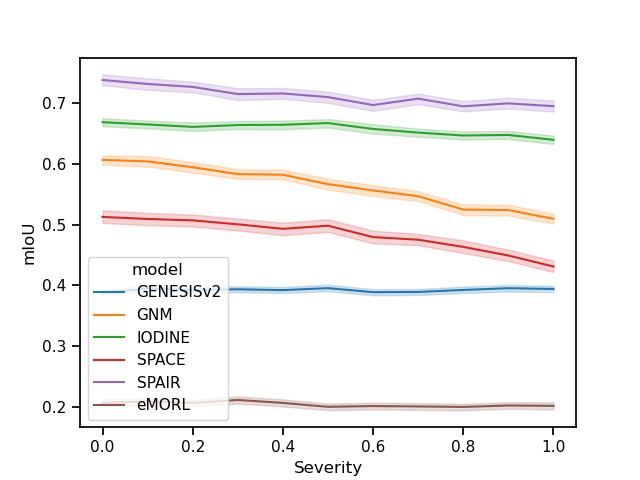} \\
    \includegraphics[width=\linewidth]{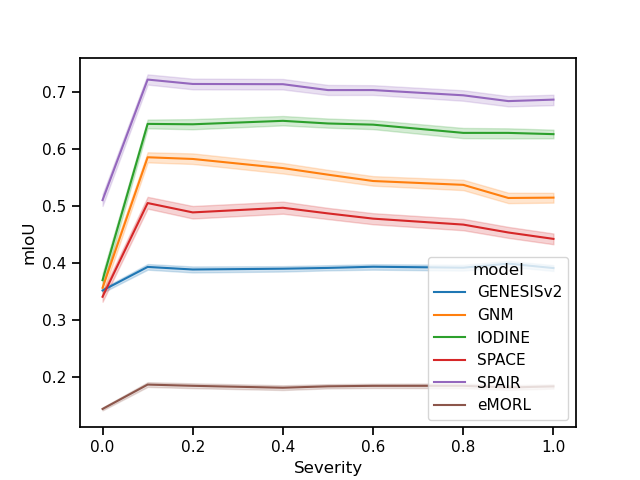}
    \caption{Blur corruption}
    \label{fig:blur}
\end{figure}

\begin{figure}[h!]
    \centering
    \includegraphics[width=\linewidth]{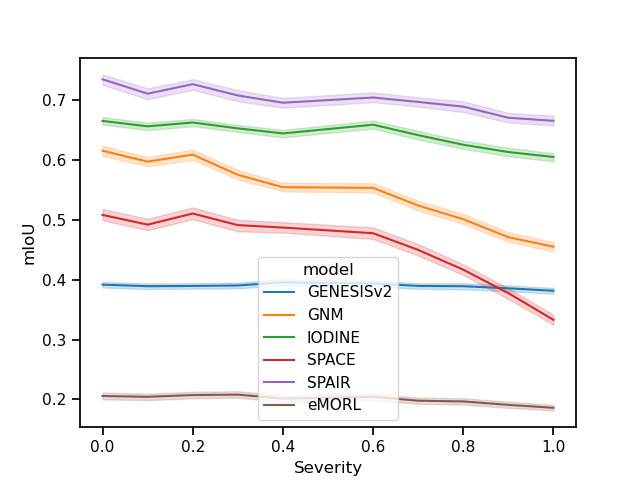} \\
    \includegraphics[width=\linewidth]{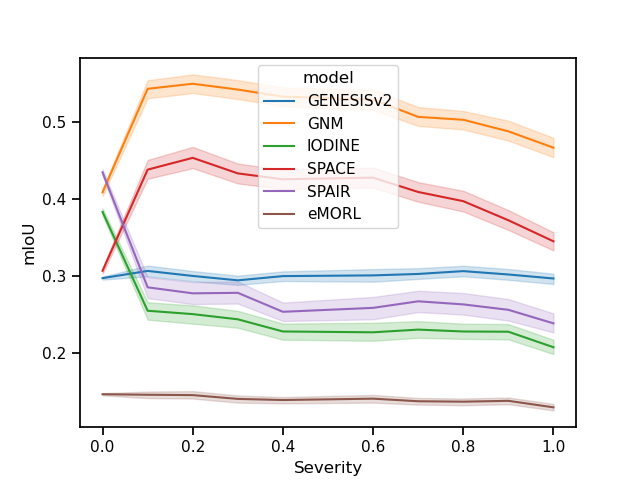} 
    \caption{Defocus Blur}
    \label{fig:defocus}
\end{figure}

\begin{figure}[h!]
    \centering
    \includegraphics[width=\linewidth]{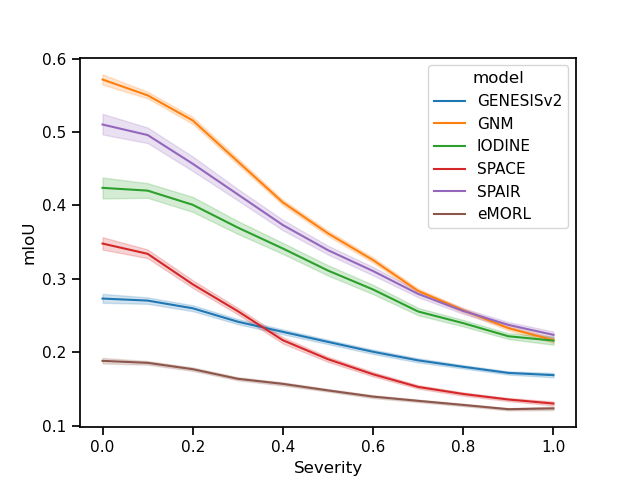} \\
    \includegraphics[width=\linewidth]{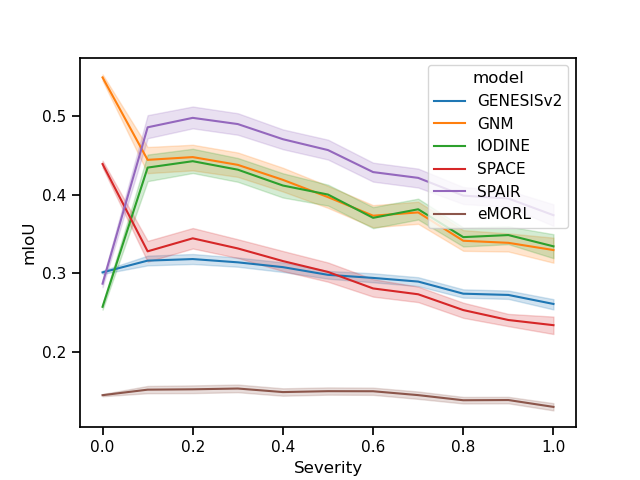}
    \caption{Displacement/motion blur}
    \label{fig:displacement}
\end{figure}

\begin{figure}[h!]
    \centering
    \includegraphics[width=\linewidth]{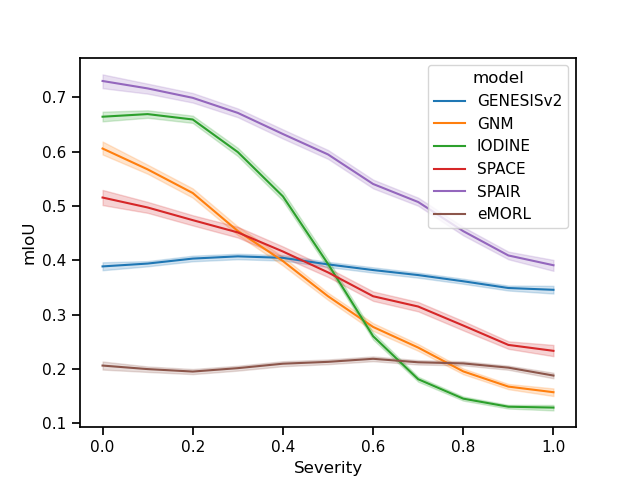} \\
    \includegraphics[width=\linewidth]{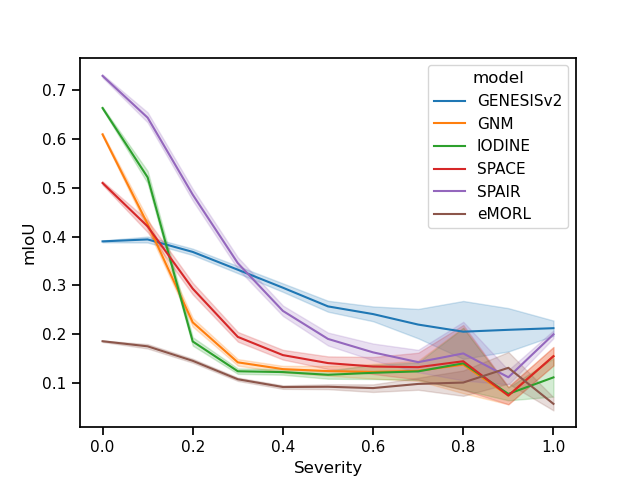}
    \caption{Clouds}
    \label{fig:clouds}
\end{figure}

\begin{figure}[h!]
    \centering
    \includegraphics[width=\linewidth]{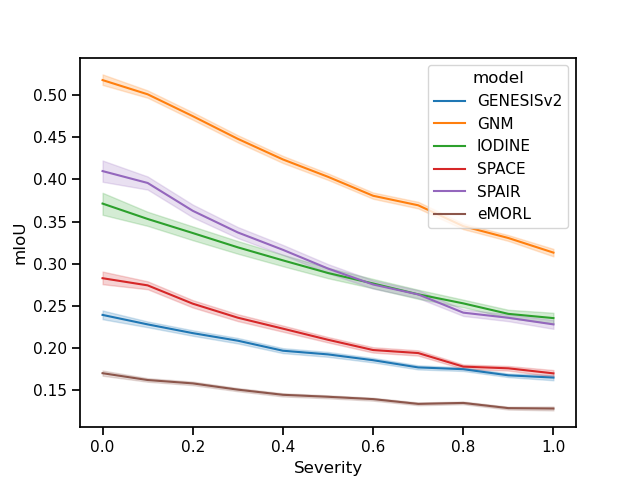} \\
    \includegraphics[width=\linewidth]{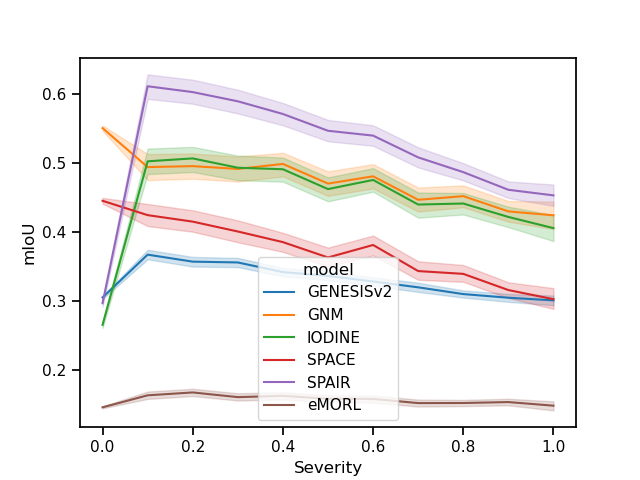} 
    \caption{Lens Distortion}
    \label{fig:lens}
\end{figure}

\begin{figure}[h!]
    \centering
    \includegraphics[width=\linewidth]{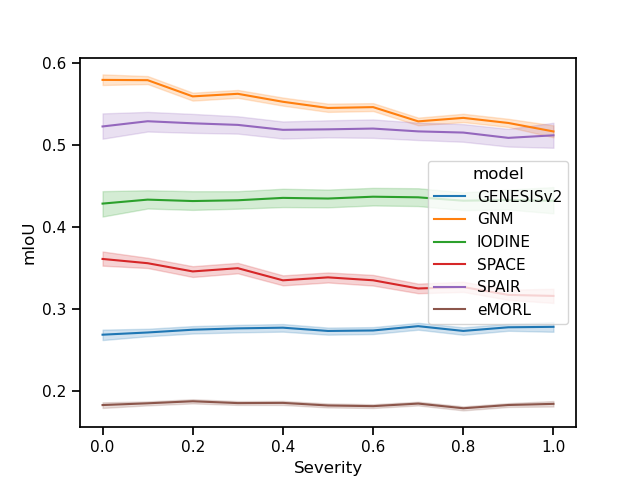} \\
    \includegraphics[width=\linewidth]{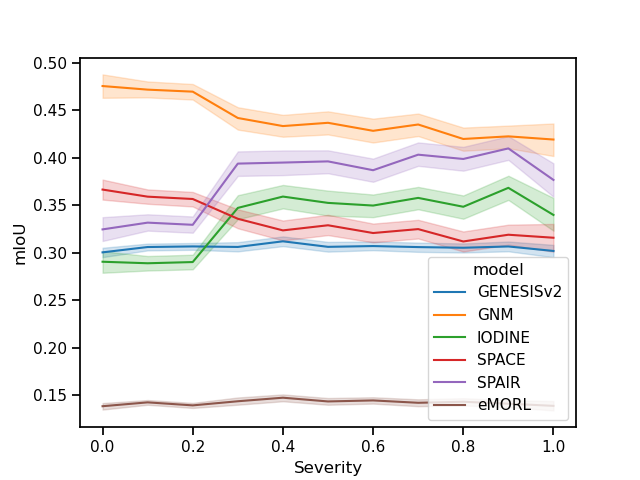} 
    \caption{White noise corruption}
    \label{fig:uniform}
\end{figure}

\end{document}